\documentclass{article}

\PassOptionsToPackage{numbers, compress}{natbib}

 \usepackage[dblblindworkshop, final]{neurips_2025_arxiv}
 \workshoptitle{AI for Non-Human Animal Communication}

\usepackage[utf8]{inputenc} %
\usepackage[T1]{fontenc}    %
\usepackage{hyperref}       %
\usepackage{url}            %
\usepackage{booktabs}       %
\usepackage{amsfonts}       %
\usepackage{nicefrac}       %
\usepackage{microtype}      %
\usepackage{xcolor}         %
\usepackage{graphicx}
\usepackage{mdframed}
\usepackage[most]{tcolorbox}
\tcbuselibrary{listings}
\usepackage{multicol}
\usepackage[nameinlink,noabbrev]{cleveref}
\usepackage{cancel}
\usepackage{wrapfig}
\usepackage{fontawesome5} %
\usepackage{academicons}  %

\newtcolorbox{promptbox}[1]{%
  enhanced,
  width=\linewidth,
  colback=blue!3,
  colframe=blue!50!black,
  colbacktitle=blue!10,
  coltitle=black,
  boxrule=0.6pt,
  arc=3pt,
  outer arc=3pt,
  left=3pt,right=3pt,top=3pt,bottom=3pt, %
  fontupper=\ttfamily\scriptsize,        %
  title=\scriptsize\bfseries #1,         %
  before skip=2pt,
  after skip=2pt
}

\newtcolorbox{answerbox}[1]{%
  enhanced,
  width=\linewidth,
  colback=gray!3,
  colframe=gray!50!black,
  boxrule=0.4pt,
  arc=3pt,
  outer arc=3pt,
  left=3pt,right=3pt,top=3pt,bottom=3pt,
  fontupper=\ttfamily\scriptsize,
  title=\scriptsize\bfseries #1
}

\usepackage{subcaption}
\usepackage[most]{tcolorbox}
\tcbset{
  enhanced,
  colback=blue!3,
  colframe=blue!35!black,
  colbacktitle=blue!10,
  coltitle=black,
  boxrule=0.6pt,
  arc=4pt, outer arc=4pt,
  left=6pt, right=6pt, top=6pt, bottom=6pt,
  title style={font=\bfseries\small\scshape},
}

\usepackage[dvipsnames]{xcolor}
\usepackage{hyperref}

\hypersetup{
    colorlinks,
    linkcolor=RoyalBlue,%
    citecolor=RoyalBlue, %
    urlcolor=RoyalBlue, %
}

\usepackage{linguex}

\newmdenv[
  backgroundcolor=yellow!20,
  linecolor=yellow!50!black,
  skipabove=1em,
  skipbelow=1em
]{draftnotes}

\tcbset{
  colback=white,
  colframe=black!15,
  boxrule=0.5pt,
  arc=2pt,
  left=6pt,right=6pt,top=4pt,bottom=4pt,
  fontupper=\footnotesize\ttfamily
}
\newcommand{\good}[1]{\textcolor{green!60!black}{#1}}
\newcommand{\bad}[1]{\textcolor{red!70!black}{#1}}
\newcommand{\llama}{\model{Llama-3.1-8B-Instruct}\xspace}
\newcommand{\naturelm}{\model{NatureLM-audio}\xspace}

\ifdefined\isNeurIPS
  \title{Model Merging Enables In-Context Learning for Bioacoustics Foundation Models}
\else
  \title{Model Merging Improves Zero-Shot Generalization in Bioacoustic Foundation Models}
\fi

\newcommand{\samethanks}[1][\value{footnote}]{\footnotemark[#1]}

\author{
Davide Marincione\textsuperscript{1,}\thanks{Equal technical contribution} \And
Donato Crisostomi\textsuperscript{1} \And
Roberto Dessi\textsuperscript{2} \AND
Emanuele Rodolà\textsuperscript{1} \And
Emanuele Rossi\textsuperscript{1,}\samethanks \AND
\mbox{}\\[-2em] %
\textsuperscript{1}Department of Computer Science, Sapienza University of Rome \quad
\textsuperscript{2}Not Diamond, San Francisco, USA
}

\newcommand{\dataset}[1]{\textsc{#1}}
\newcommand{\model}[1]{\textsc{#1}}

\begin{document}

\maketitle
\begin{abstract}

Foundation models capable of generalizing across species and tasks represent a promising new frontier in bioacoustics, with \naturelm being one of the most prominent examples. While its domain-specific fine-tuning yields strong performance on bioacoustic benchmarks, we observe that it also introduces trade-offs in instruction-following flexibility. For instance, \naturelm{} achieves high accuracy when prompted for either the common or scientific name individually, but its accuracy drops significantly when both are requested in a single prompt.  
We address this by applying a simple model merging strategy that interpolates \naturelm{} with its base language model, recovering instruction-following capabilities with minimal loss of domain expertise. Finally, we show that the merged model exhibits markedly stronger zero-shot generalization, achieving over a 200\% relative improvement and setting a new state-of-the-art in closed-set zero-shot classification of unseen species. 
\begin{center}
\vspace{-0.3em}
\href{https://github.com/gladia-research-group/model-merging-NatureLM-audio}{\faGithub\ \texttt{gladia-research-group/model-merging-NatureLM-audio}}
\vspace{-1em}
\end{center}

\end{abstract}

\section{Introduction}

\begin{wrapfigure}[15]{r}{0.45\textwidth}
  \centering
  \vspace{-15pt}
  \includegraphics[width=\linewidth]{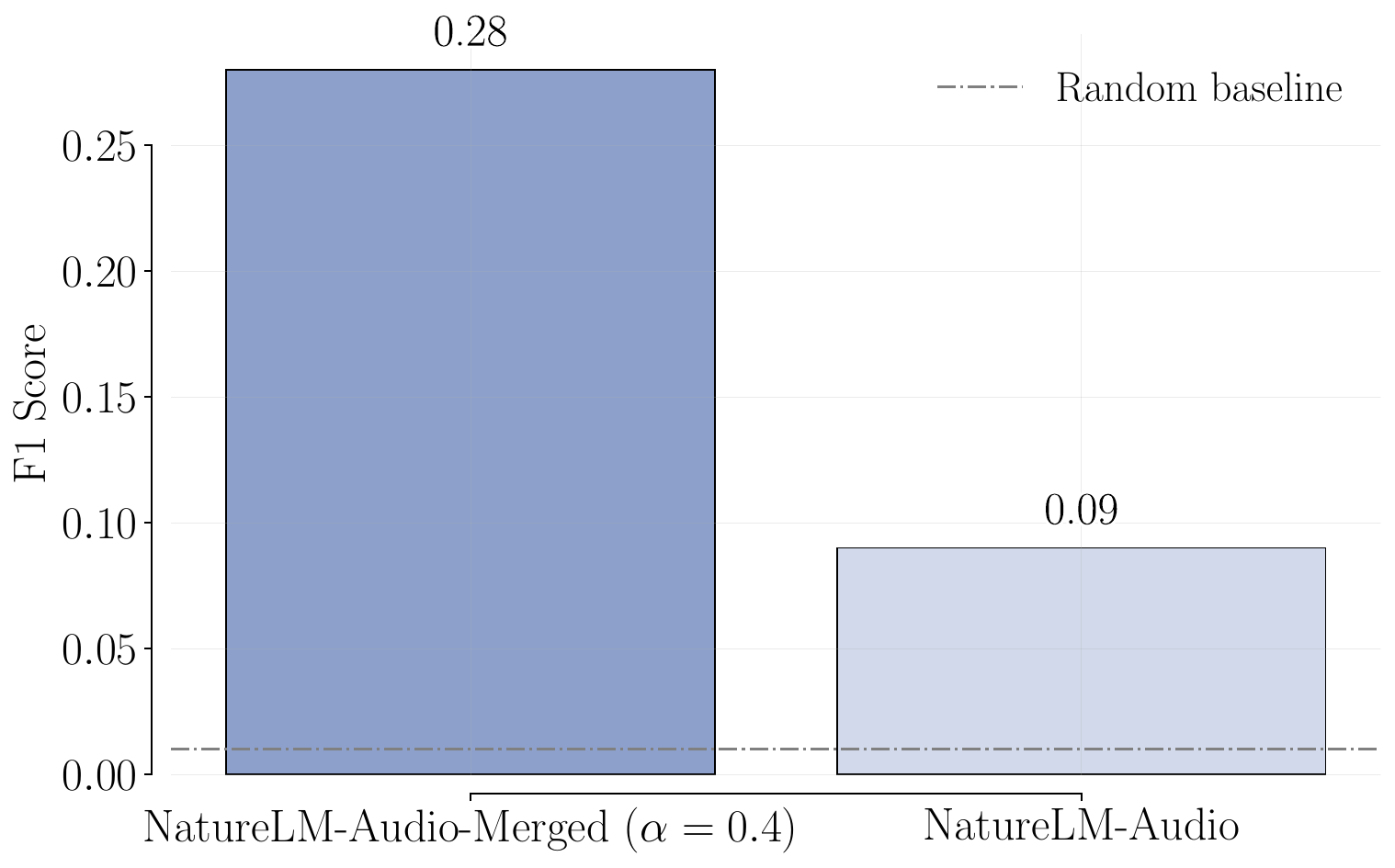}
  \caption{
    Model merging leads to a \textit{200\% relative improvement} over \naturelm in zero-shot classification of unseen species, setting a new state-of-the-art. 
  }
  \label{fig:unseen_cmn_family_barplot}
\end{wrapfigure}

Bioacoustics, the study of sound production, transmission, and perception in animals, is a critical tool for understanding biodiversity, monitoring ecosystems, and informing conservation efforts~\citep{marler2004nature, laiolo2010emerging, marques2013estimating}. Recent advances in machine learning (ML) have transformed the field, enabling automated detection, classification, and analysis of acoustic events at unprecedented scales~\citep{stowell2022computational}.  

Early work in ML for bioacoustics typically relied on \emph{species-specific models}, trained and optimized for a single species and task~\citep{Aide2013PeerJ_e103}. However, as in other domains of ML, there is now a shift towards \emph{general-purpose foundation models} that can support a broad range of downstream species and/or tasks with minimal retraining~\citep{kahl2021birdnet, ghani2023global, hagiwara2023aves, rauch2025maskedautoencoderslistenbirds, robinson2023transferablemodelsbioacousticshuman, vanmerriënboer2025perch20bitternlesson}. One of the most prominent examples of this trend is \model{NatureLM-audio}~\citep{robinson2025naturelm}, the first bioacoustics audio--language model, designed for zero-shot generalization to unseen tasks via text-based prompting.  

In this paper, we examine the capabilities of \model{NatureLM-audio} as a \emph{general} foundation model for bioacoustics. Despite its strong performance on tasks and prompts closely matching its training distribution, we find that its intense domain-specific fine-tuning has led to a severe reduction in the \emph{instruction-following capabilities} of its base model (\llama), a trade-off commonly observed in other specialized models~\citep{zhai2023investigating} and limiting its ability to generalize in zero-shot settings. We show that \emph{model merging} with the base model can help restore these capabilities, achieving a balance between domain-specific knowledge and general instruction-following. Finally, we show that this approach sets a new \emph{state-of-the-art in zero-shot classification of unseen species}, achieving over a 200\% relative improvement compared to \naturelm~(\cref{fig:unseen_cmn_family_barplot}).

\section{Problem Analysis} \label{sec:problem_analysis}

\begin{figure}[t]
  \centering
  \begin{subfigure}[t]{0.47\textwidth}
    \vspace{0pt}%
    \begin{promptbox}{Common Name Prompt}
What is the common name for the focal species in the audio?
    \end{promptbox}\vspace{4pt}
    \begin{promptbox}{Scientific Name Prompt}
What is the scientific name for the focal species in the audio?
    \end{promptbox}\vspace{4pt}
    \begin{promptbox}{Combined Prompt}
Identify the focal species in the audio and provide its scientific name,
followed by a colon and its common name.
    \end{promptbox}
  \end{subfigure}\hfill
  \begin{subfigure}[t]{0.49\textwidth}
    \vspace{0pt}%
    \vspace{0.65cm}
    \includegraphics[width=\linewidth]{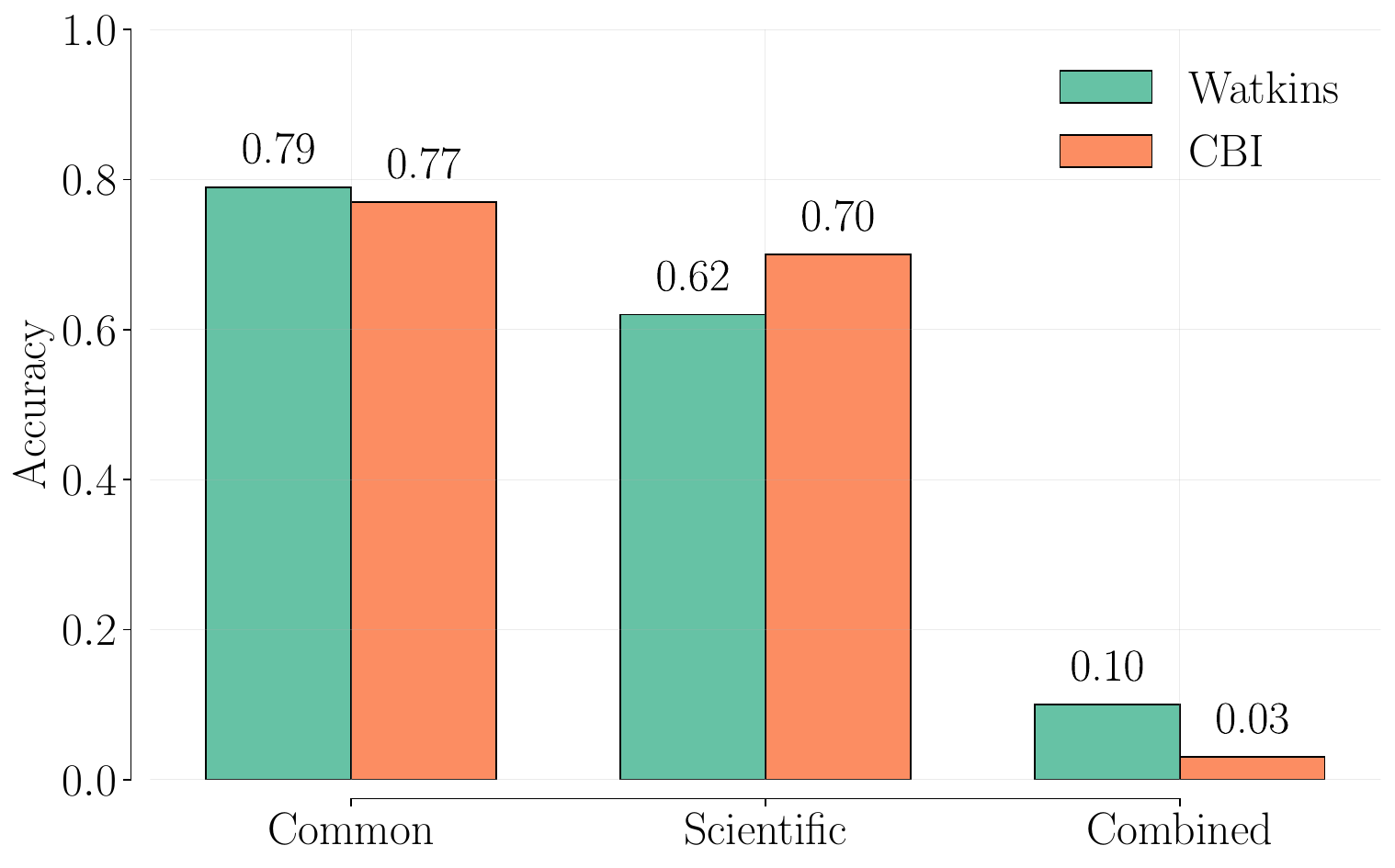}
  \end{subfigure}
  \vspace{-4pt}
  \caption{
    \naturelm classification accuracy for different prompts on \dataset{Watkins} and \dataset{CBI}.
  }
  \label{fig:scientific_common_combined}
\end{figure}

\begin{figure}[t]
\vspace{-3pt}
\centering
\begin{subfigure}[t]{0.49\textwidth}
\begin{answerbox}{id aecd4535-ebef-438d-811b-1ffb7be5c22e}
\textbf{GT:} Odobenus rosmarus: Walrus

\vspace{2pt}
Common: \good{Walrus} \\
Scientific: \good{Odobenus rosmarus} \\
Combined: \bad{Odobenus rosmarus - male courtship ...}
\end{answerbox}

\begin{answerbox}{id 185c974a-d905-475c-8fc0-0cbab1e383b9}
\textbf{GT:} Eubalaena australis: Southern Right Whale

\vspace{2pt}
Common: \bad{Fin- Finback Whale} \\
Scientific: \bad{Balaenoptera physalus} \\
Combined: \bad{Balaenoptera physalus: 52 Hz Pulses}
\end{answerbox}
\end{subfigure}%
\hfill
\begin{subfigure}[t]{0.49\textwidth}
\begin{answerbox}{id 4cbdd583-23c6-408d-aea5-3719dc6d9654}
\textbf{GT:} Lagenodelphis hosei: Frasers Dolphin

\vspace{2pt}
Common: \good{Fraser's Dolphin} \\
Scientific: \good{Lagenodelphis hosei} \\
Combined: \bad{Lagenodelphis hosei}
\end{answerbox}

\begin{answerbox}{id 4186fc55-0ac2-4f44-9026-009f239fcf96}
\textbf{GT:} Stenella clymene: Clymene Dolphin

\vspace{2pt}
Common: \good{Clymene Dolphin} \\
Scientific: \good{Stenella clymene} \\
Combined: \good{Stenella clymene: Clymene Dolphin}
\end{answerbox}
\end{subfigure}

\caption{Example model predictions for the common name, scientific name, and combined-name prompts, compared to ground truth. Correct predictions in \good{green}, incorrect in \bad{red}.}
\label{fig:prompt_examples}
\end{figure}

\model{NatureLM-audio} is a LoRA~\citep{hu2022lora} fine-tuning of \model{Llama-3.1-8B-Instruct}~\citep{grattafiori2024llama} on $\sim$2M steps of audio–text pairs, predominantly bioacoustics but also music and human sounds.
While the original evaluation shows that the model follows training-like instructions well, such as predicting either the common or the scientific name of the focal species in an audio in isolation, we find that requesting both in a single prompt often leads to a large drop in accuracy. \Cref{fig:scientific_common_combined} shows the prompts and corresponding accuracies on \dataset{Watkins} and \dataset{CBI}, two species-classification datasets from the \dataset{BEANS-Zero} benchmark~\citep{hagiwara2023beans} covering marine mammals and birds, respectively. On both datasets, the model performs better on common names than on scientific names, yet achieves high accuracy (60–80\%) in both cases. However, when prompted for both names jointly, accuracy falls to 6–12\%.

The examples in~\cref{fig:prompt_examples} illustrate typical failure modes.  In the top left, the model outputs correct common and scientific names individually, but under the combined prompt it drifts into behavioural description (“male courtship behavior”), possibly reflecting its exposure to captioning-style data during training~\citep{robinson2025naturelm}. In the bottom left, it misidentifies the species in all cases, yet common and scientific predictions remain mutually consistent; in the combined case it again appends unrelated context (“52~Hz pulses”). In the top right, correct individual predictions degrade to only the scientific name under the combined prompt. In the bottom right, all three prompts succeed.

We additionally experiment with the \dataset{zf-indiv} dataset originally used in~\citep{robinson2025naturelm} to evaluate zero-shot task generalization (see~\cref{sec:zf-indiv}) and observe a similar pattern: \model{NatureLM-audio} shows reduced robustness to even mild prompt variations. This behaviour is consistent with the effects of extensive domain-specific fine-tuning observed in other specialized LLMs, where overfitting to training prompt formats can narrow instruction-following flexibility and limit generalization~\citep{zhai2023investigating}.

\section{Method}
\Cref{sec:problem_analysis} shows that \naturelm has lost its instruction following capabilities in favor of task-specific ones acquired during fine-tuning. We recover these ones through model merging.

\paragraph{Model Merging}
Model merging aims to ensemble different models without incurring in additional inference or storage costs~\cite{git-rebasin,garipov_loss_2018, model-soups, cycle-consistent}. While the non-linear nature of neural networks prevents from taking the weighted average of the models in general~\cite{wortsman_robust_2022}, this aggregation is well behaved when the two models exhibit linear mode connectivity~\cite{linear-mode-connectivity}, \emph{i.e.} can be connected via a linear path over which the loss does not significantly increase. In this case, the merged model $\mathbf{\Theta}^{(\text{merge})}$ can be obtained from the endpoint models $\mathbf{\Theta}^{(1)} , \mathbf{\Theta}^{(2)}$ simply as $\mathbf{\Theta}^{(\text{merge})} = (1 - \alpha) \mathbf{\Theta}^{(1)} + \alpha \mathbf{\Theta}^{(2)}$, 
where $\alpha \in [0, 1]$ is a scaling parameter controlling the contribution of each model. Consistent with previous findings \citep{wortsman_robust_2022, neyshabur2020being, linear-mode-connectivity}, we observe that linear interpolation remains effective along the fine-tuning trajectory, suggesting that linear mode connectivity holds when (part of) the optimization path is shared.

\paragraph{Merging \naturelm with its base model}
We merge \llama with its fine-tuning \naturelm to combine the instruction following abilities of the former and the task-specific performance of the latter.
In particular, being \model{NatureLM-audio} a LoRa~\cite{hu2022lora} fine-tuning, linearly interpolating between the base and the fine-tuned is equivalent to changing the multiplicative factor $\alpha$ in LoRa: Given the weight matrix $\mathbf{W}_{\text{base}}$ 
of the base model for some layer, LoRA updates it as $\mathbf{W}_{\text{ft}} = \mathbf{W}_{\text{base}} + \mathbf{A}\mathbf{B}$, where $\mathbf{A}$ and $\mathbf{B}$ are two low-rank learnable matrices; thus 
\begin{align}
    (1 - \alpha)\,\mathbf{W}_{\text{base}} + \alpha\,\mathbf{W}_{\text{ft}} 
    &= (1 - \alpha)\,\mathbf{W}_{\text{base}} + \alpha\left( \mathbf{W}_{\text{base}} + \mathbf{A}\mathbf{B} \right) \\
    &= \mathbf{W}_{\text{base}} - \cancel{\alpha\,\mathbf{W}_{\text{base}}} + \cancel{\alpha\,\mathbf{W}_{\text{base}}} + \alpha\,\mathbf{A}\mathbf{B} = \mathbf{W}_{\text{base}} + \alpha\,\mathbf{A}\mathbf{B}.
\end{align}
This shows that we can interpolate between the base and the fine-tuned model simply by varying ~$\alpha$.

\section{Results}

\paragraph{Combined Instruction-Following Task}
\begin{wrapfigure}[17]{r}{0.45\textwidth} %
    \centering
    \vspace{-32pt} %
    \includegraphics[width=0.40\textwidth]{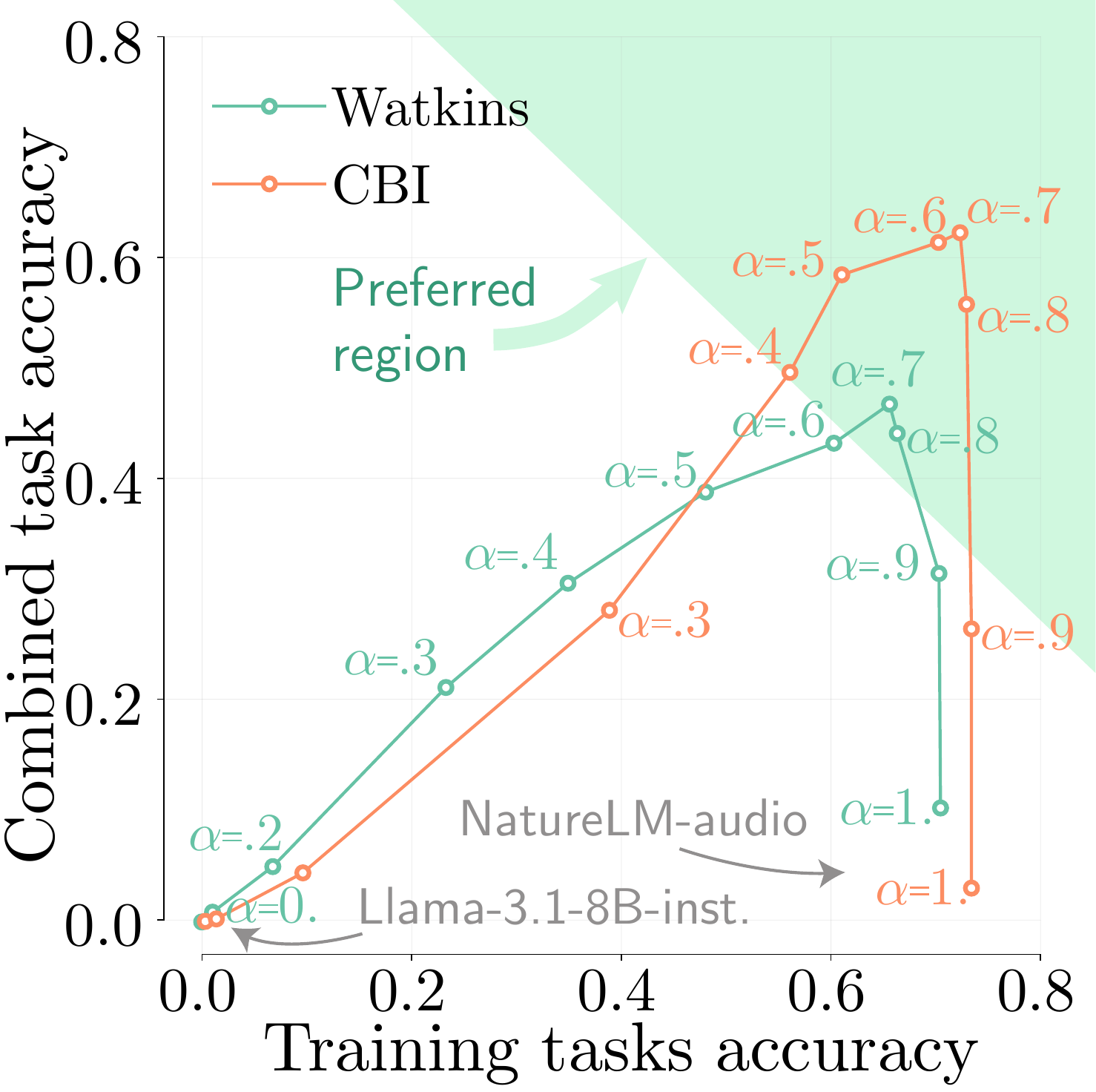}
    \vspace{-5pt}
    \caption{Accuracy on the combined prompt (y-axis) from~\cref{fig:scientific_common_combined} 
    versus the mean accuracy on the individual common- and scientific-name prompts (x-axis) when varying ~$\alpha$. }
    \label{fig:scientific_common_combined_grid}
    \vspace{-25pt}
\end{wrapfigure}
We evaluate the merged model over a range of interpolation coefficients~$\alpha$, using the three prompt variants in~\cref{fig:scientific_common_combined}. The $y$-axis in~\cref{fig:scientific_common_combined_grid} reports the accuracy on the \emph{combined} prompt, while the $x$-axis shows the mean accuracy on the \emph{training-like} prompts (common name and scientific name individually).
For the combined prompt, intermediate interpolation values substantially outperform both extremes. Specifically, rescaling from $\alpha = 1$ (\model{NatureLM-audio}) to $\alpha \approx 0.7$ increases combined-task accuracy from $6\% \rightarrow 45\%$ on Watkins and $12\% \rightarrow 63\%$ on CBI, reflecting a restoration of instruction-following capabilities degraded in the fine-tuned model. However, setting $\alpha$ too low ($\alpha < 0.5$) sharply reduces accuracy on combined prompts due to a loss of domain-specific audio knowledge from the fine-tuning stage.

The observed behaviour highlights $\alpha$ as a controllable \emph{capability trade-off parameter}. At $\alpha = 1$, the model fully retains its domain adaptation but suffers in compositional instruction following. At $\alpha = 0$, it maximizes general instruction-following behaviour inherited from the base model, but discards most bioacoustic specialization. The monotonic decline in $x$-axis accuracy with decreasing $\alpha$ further confirms that domain-task performance and instruction-following ability are in tension.

\paragraph{Classification of Unseen Species}

We next evaluate the merged model on the task of \textit{zero-shot classification of previously unseen species}, using the \dataset{unseen-family-cmn} split of the \dataset{BEANS-Zero} benchmark~\citep{robinson2025naturelm}. This split, comprising 425 samples from 40 classes, enforces that no species from the same taxonomic family appear in both training and test sets. The task is to predict each focal species’ \textit{common name} without any fine-tuning.
Unlike the open-vocabulary formulation of \dataset{BEANS-Zero}, we adopt a \textit{closed-set} setup where all possible 40 target labels are provided in the prompt, allowing us to assess adherence to label constraints. This setting also mirrors realistic field conditions where the set of possible species is geographically bounded.

\begin{wrapfigure}[15]{r}{0.42\textwidth}
    \centering
    \vspace{-14pt} %
    \includegraphics[width=\linewidth]{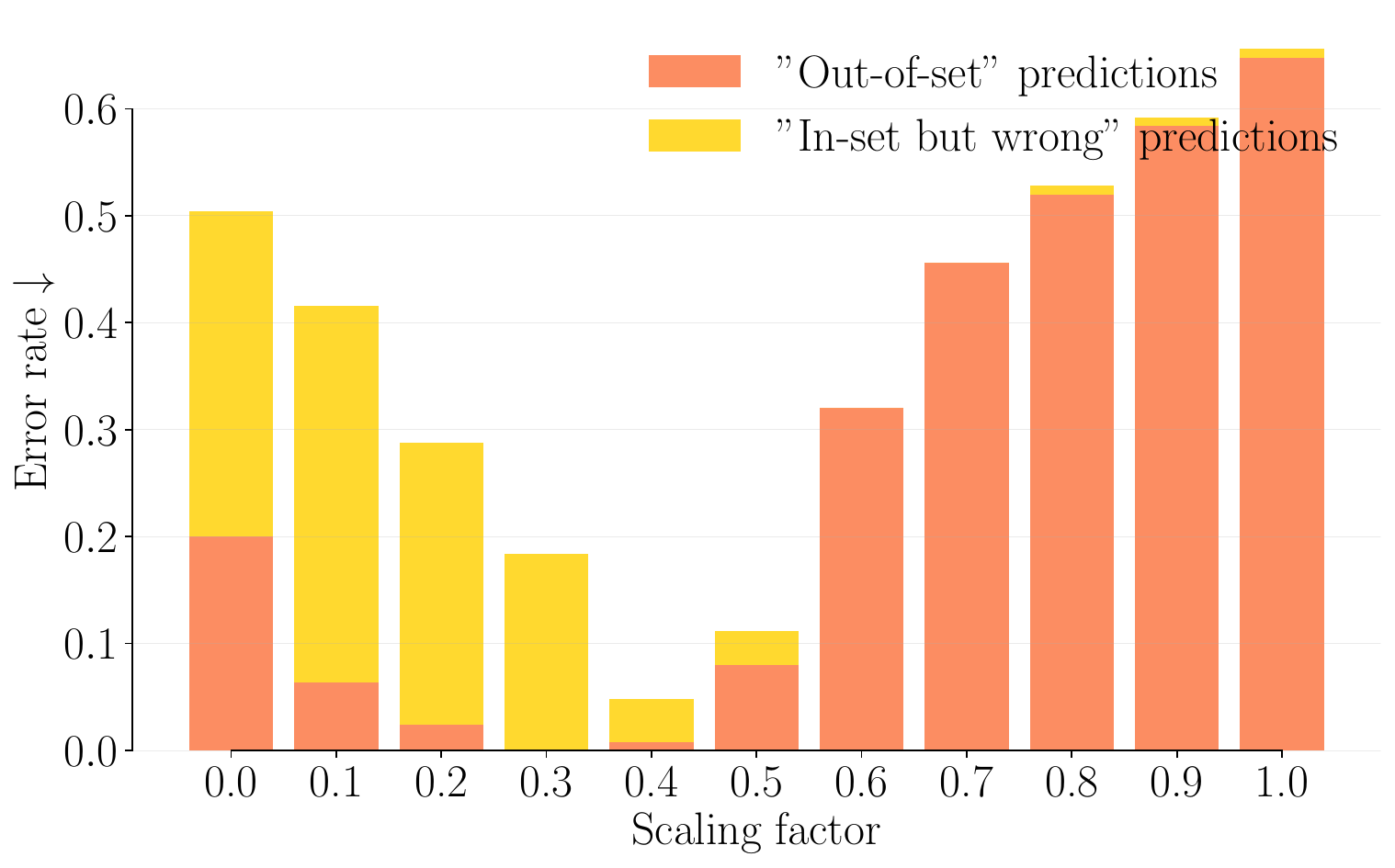}
    \caption{Error breakdown (lower is better) on the binary subset of \dataset{unseen-family-cmn} as a function of the merging coefficient $\alpha$.}
    \label{fig:unseen_cmn_family}
\end{wrapfigure}

Quantitative results (\cref{fig:unseen_cmn_family_barplot}) show that the merged model achieves a \textit{200\% relative improvement} over \naturelm (F1 = 0.28 vs. 0.09), demonstrating markedly stronger generalization to unseen taxa and setting a new state-of-the-art. To better understand this gain, we analyze a simplified binary version of the task where the model has to predict only among the two most popular classes (125 samples,~\cref{fig:unseen_cmn_family}). For $\alpha \in [0.6, 1.0]$, the model rarely confuses valid classes but frequently produces out-of-set predictions, signaling strong discrimination but weak prompt adherence. As $\alpha$ decreases, these invalid outputs vanish, reaching an optimal trade-off around $\alpha = 0.4$, beyond which in-set confusions and abstentions slightly increase.
The results suggest that the \naturelm audio encoder already provides robust species-level representations, and that its main bottleneck lies in the language model’s capacity to follow prompts. Model merging effectively mitigates this limitation and greatly improves \textit{closed-set zero-shot classification} by enhancing adherence to the prompt constraints.

\section{Related Work}

\paragraph{Catastrophic forgetting in multi-modal fine-tuning}
Catastrophic forgetting is a well-known challenge when fine-tuning large language models~\citep{Shi2024ContinualLOA}. A common training-time mitigation is to freeze the LLM and update only the projection layer that maps visual or audio features into the text embedding space, often with fewer fine-tuning steps~\citep{zhai2023investigating} or using PEFT methods~\citep{conf/acl/ZhaoWHZQZYXC24, panda2024lottery}. In contrast, post-training approaches aim to restore forgotten skills in already fine-tuned models~\citep{modeltailor2024, pmlr-v202-panigrahi23a}.

\paragraph{Model merging} Model merging provides an efficient alternative to ensembling, producing a single model combining multiple models' capabilities without increasing inference cost. Early work, inspired by linear mode connectivity~\citep{linear-mode-connectivity,garipov_loss_2018,mirzadeh_linear_2020,Entezari2021-me}, focused on aligning independently trained models, often by solving a neuron permutation problem~\citep{git-rebasin,repair,cycle-consistent,model-fusion,zip-it,rebasin-implicit-sinkhorn}. Closer to our work, \citet{wortsman_robust_2022} produce robust fine-tuned models by linearly interpolating them with their base model, while \citet{ilharco_patching_2022} use interpolations to improve specific tasks without waiving others.

\section{Conclusions}
We investigated the instruction-following limitations of \naturelm and found that even small changes in prompt structure can significantly degrade performance, reducing its utility as a general-purpose model. To address this, we applied a lightweight model-merging strategy that interpolates the fine-tuned \naturelm with its base model. Intermediate interpolation weights restore much of the lost instruction-following capability while preserving most domain-specific accuracy. This recovery further improves zero-shot classification of unseen species, setting a new state-of-the-art. In our experiments, $\alpha \in [0.4, 0.6]$ provided a strong balance between instruction following and domain expertise, though the optimal value remains task- and dataset-dependent.

\paragraph{Limitations and Future Work}
Our current evaluation of zero-shot closed-set classification is limited in scope and we plan to extend it to multiple datasets in the future.
Convex weight interpolation may not be optimal, and we intend to explore alternative strategies for restoring instruction-following abilities, including more advanced model-merging methods~\citep{sakana, mencattini2025merge, TSV, Iso-C, stoicamodel}) and activation-steering techniques~\citep{arditi2024refusal, stolfoimproving}.

\appendix

\newpage

\section*{Acknowledgments}
We thank Emmanuel Chemla for his helpful feedback. This work is partly supported by the MUR FIS2 grant n. FIS-2023-00942 ”NEXUS” (cup B53C25001030001), and by Sapienza University of Rome via the Seed of ERC grant ”MINT.AI” (cup B83C25001040001).

{
\small
\bibliography{references/merging, references/other}
\bibliographystyle{plainnat}
}

\section{Extended related work}
\paragraph{Foundation Models in Bioacoustics}  
Recent advances have introduced large-scale bioacoustic foundation models designed for cross-species and cross-task generalization. \model{NatureLM-audio}~\citep{robinson2025naturelm} integrates a self-supervised audio encoder with a LLaMA-based language decoder. \model{BioLingual}~\citep{robinson2023transferablemodelsbioacousticshuman} adapts CLAP-style audio–text contrastive learning to bioacoustics, while audio-only models such as \model{BirdMAE}~\citep{rauch2025maskedautoencoderslistenbirds}, \model{AVES}~\citep{hagiwara2023aves}  and \model{Perch 2.0}~\citep{vanmerriënboer2025perch20bitternlesson} pretrain large models on extensive birdsong or multi-taxa datasets to produce broadly transferable acoustic features. Although these models surpass species-specific baselines, they remain susceptible to domain shifts and catastrophic forgetting, which limits their robustness in real-world deployments.

\paragraph{Mode connectivity and model merging}
Mode connectivity investigates the weight configurations that define local minima. \citet{linear-mode-connectivity} examines the linear mode connectivity of models trained for only a few epochs from the same initialization, linking this phenomenon to the lottery ticket hypothesis. Relaxing the shared-initialization requirement, \citet{Entezari2021-me} argues that, after resolving neuron permutations~\cite{navon2023equivariant,horoi_harmony_2024}, all trained models may reside in a single connected basin.
Model merging pursues a different goal: combining multiple models into one that inherits their capabilities without the cost and complexity of ensembling. In this direction, \citet{model-fusion} introduced an optimal-transport–based weight-matching method, while \texttt{Git Re-Basin}~\cite{git-rebasin} proposes optimizing a linear assignment problem (LAP) for each layer. More recently, \texttt{REPAIR}~\cite{repair} shows that substantial gains in the performance of interpolated models can come from renormalizing activations, while $C^2M^3$~\citep{cycle-consistent} proposes matching and merging many models jointly through cycle-consistent permutations. When the models to merge are fine-tuned from a shared backbone, task-vector-based methods are most effective~\citep{task-vectors, ties, yu2024language, matena_merging_2022, wortsman_robust_2022, davari2025model, wang2024localizing, zhou2024atmimprovingmodelmerging, TSV, ortiz2024task, mencattini2025merge, huang2024emrmerging, daheim2023model, stoicamodel, yang2024representation, tangparameter}. These involve taking the parameter-level difference between the fine-tuned model and its pretrained base, termed a task vector. Improvements can be obtained by optimizing task-vector combinations~\citep{yang2024adamerging}, mitigating sign disagreement~\cite{ties}, randomly dropping updates~\cite{yu2024language}, or applying evolutionary methods~\cite{sakana, mencattini2025merge}. Finally, techniques employing layer-wise task vectors~\cite{stoicamodel, TSV, Iso-C} obtain state-of-the-art results by leveraging layer-level structures through SVD of the parameter differences.

\section{Additional Experiments} 

\subsection{Combined Instruction Following Task}
\begin{figure}
\centering
\includegraphics[width=0.9\textwidth]{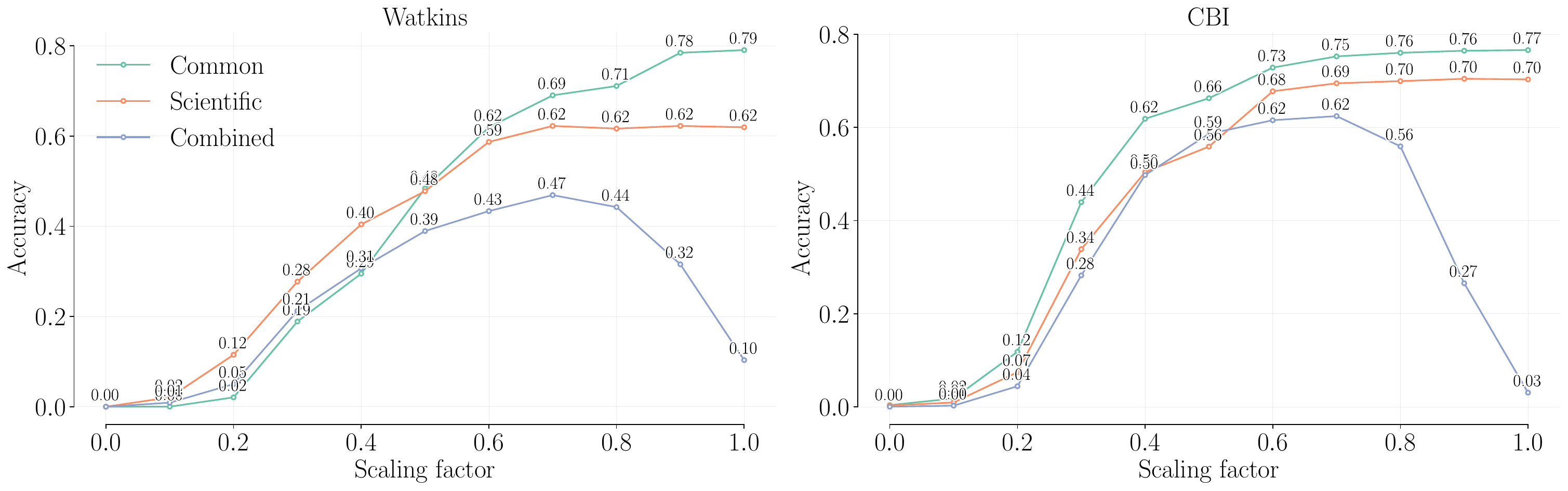}
\caption{Accuracy on the common name, scientific name, and combined prompts from~\cref{fig:scientific_common_combined}, as a function of the rescaling parameter~$\alpha$.}
\label{fig:scientific_common_combined_grid_explicit}
\end{figure}

\Cref{fig:scientific_common_combined} shows that the accuracy of the original \model{NatureLM-audio} model on the ``combined prompt'' is drastically lower than its ``common'' and ``scientific'' counterparts, and \cref{fig:scientific_common_combined_grid} shows how the accuracy of the ``combined prompt'' varies as a function of the accuracy over the other two prompts. Here, \Cref{fig:scientific_common_combined_grid_explicit} provides a more granular visualization, explicitly plotting how the accuracy of each prompt evolves with changes in the rescaling parameter~$\alpha$, thereby highlighting the trade-off between instruction-following recovery and domain-specific retention.

\subsection{Experimental Details}
\paragraph{Correct prediction estimation} Following \cite{robinson2025naturelm}, we evaluate classification accuracy by first extracting the model’s free-form output and then computing the Levenshtein distance between this output and each possible target class name (in this case, the two species' common names). The class with the smallest distance is selected, and the prediction is considered correct if this distance is less than a threshold $t$ (set to $t=5$ in our experiments).

For small $\alpha$, we often observe that the model tends to output a consistent prefix sentence before presenting the actual answer. When prompted for the common/scientific name (on any experiment) the model responds with ``\texttt{The common/scientific name for the focal species in the audio is \{species\_name\}}'', and specifically, when a scientific name is requested, the model sometimes also follows with ``\texttt{, also known as \{common\_name\_of\_species\}}''. Due to the formulaic nature of these answers, we decided to check for such cases and extract the model's actual prediction from them, before computing the above-mentioned distance computation (as otherwise, they would be flagged as incorrect).

\paragraph{Classification of Unseen Species} To mitigate position bias, we randomly permute the order of few-shot examples for each evaluation sample. This randomization is applied independently for every sample, thereby minimizing spurious correlations between class position and model predictions. As previously noted, the two species used in the experiment, Spotted Elachura (\emph{Elachura formosa}) and Dall's Porpoise (\emph{Phocoenoides dalli}), were selected for being the most frequent classes in the \dataset{unseen-family-cmn} dataset, with 73 and 53 samples respectively.

\begin{figure}[t]
  \centering
  \begin{promptbox}{Closed-Set Classification Prompt}
    What is the common name for the focal species in the audio? Output exactly one of: \{species\_list\} 
  \end{promptbox}
  \begin{subfigure}[t]{0.47\textwidth}
    \vspace{0pt}%
    \includegraphics[width=\textwidth]{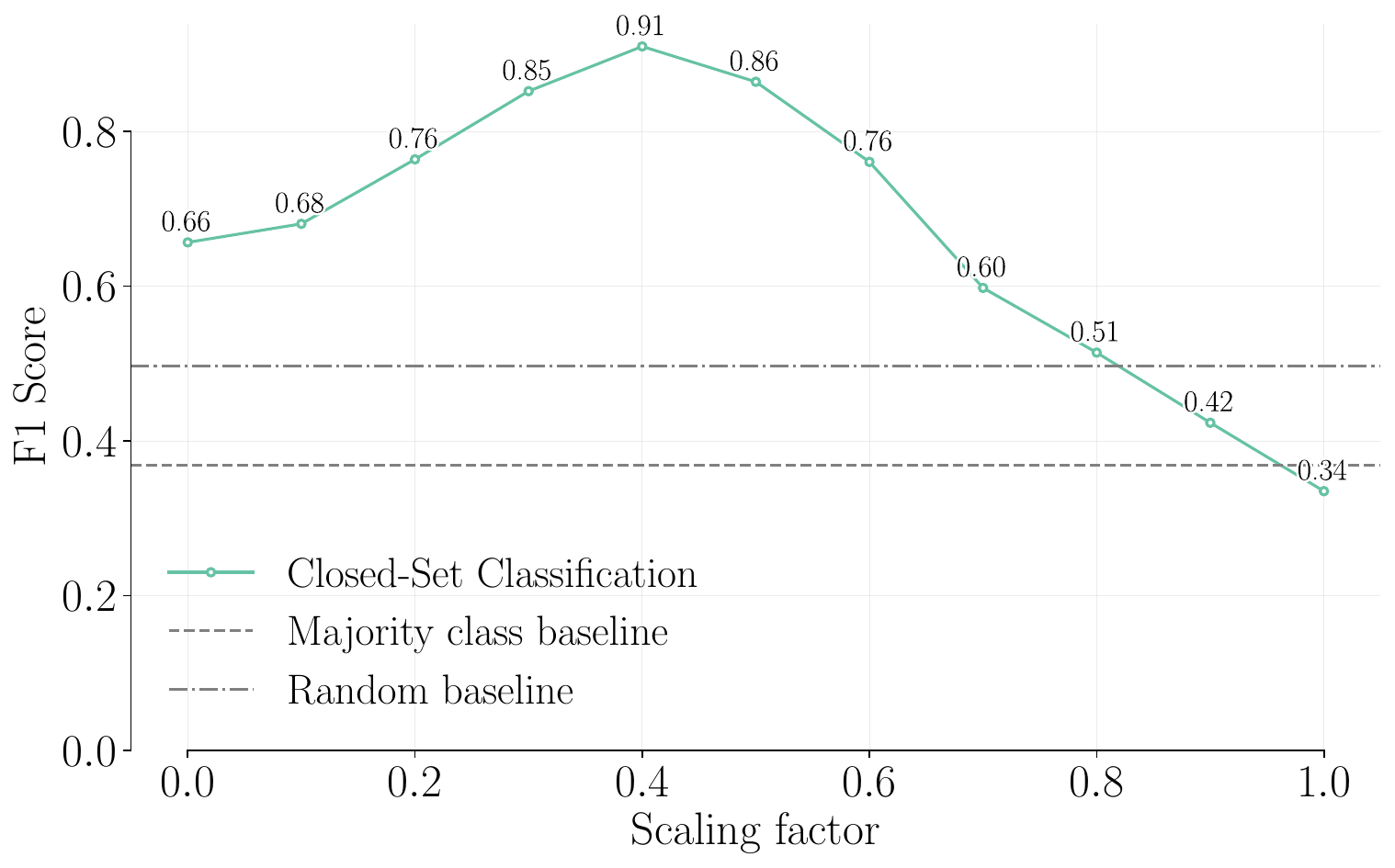}
  \end{subfigure}\hfill
  \begin{subfigure}[t]{0.49\textwidth}
    \vspace{0pt}%
    \includegraphics[width=\textwidth]{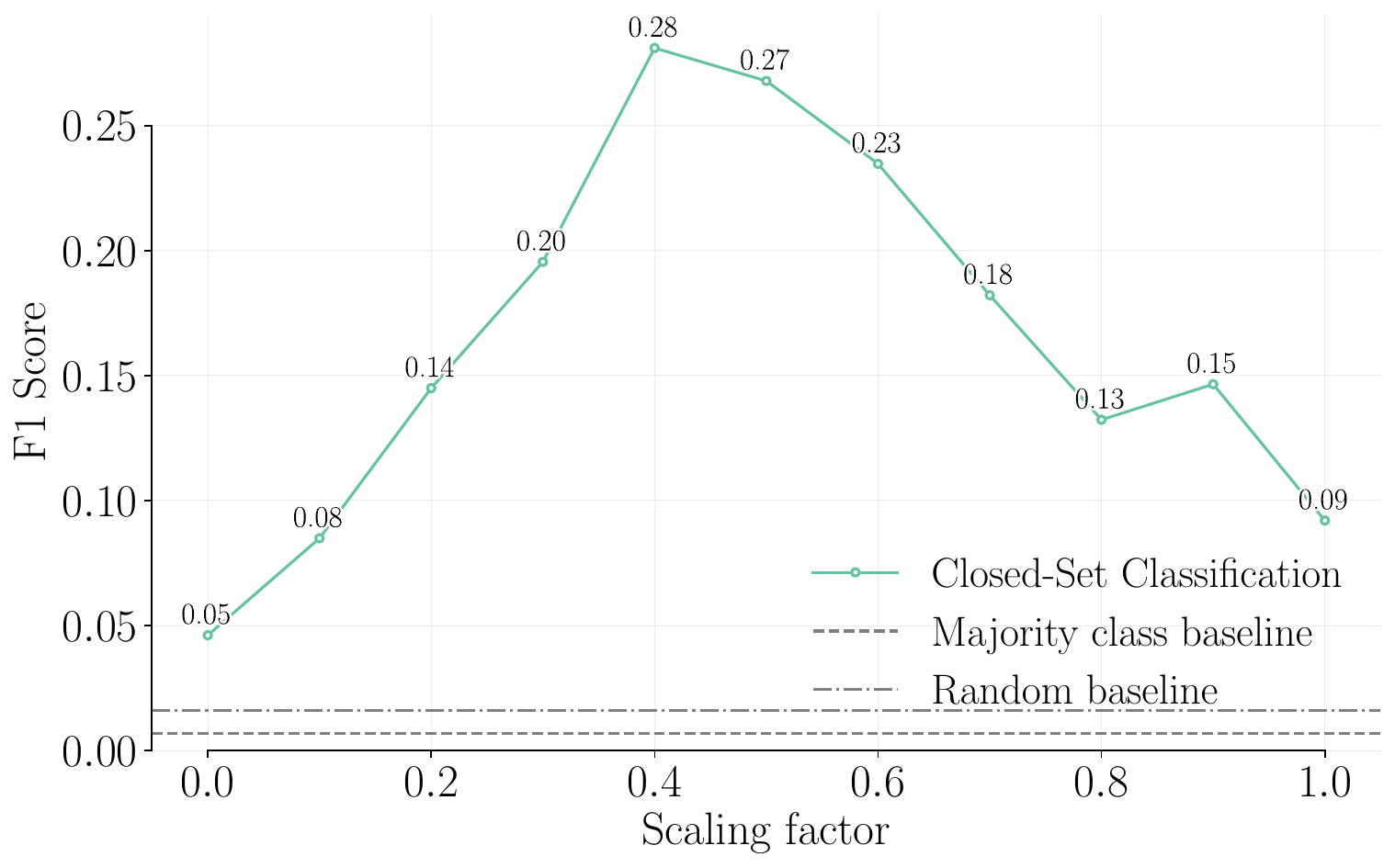}
  \end{subfigure}
  \vspace{-4pt}
  \caption{
    F1 classification score on \dataset{unseen-cmn-family} when varying $\alpha$. In the left plot, only the two most popular classes are kept for the task, while on the right plot all classes are used. Note the different scales of the plots.
  }
  \label{fig:unseen_cmn_family_appendix}
\end{figure}

\begin{figure}[t]
  \centering
  \includegraphics[width=0.5\textwidth]{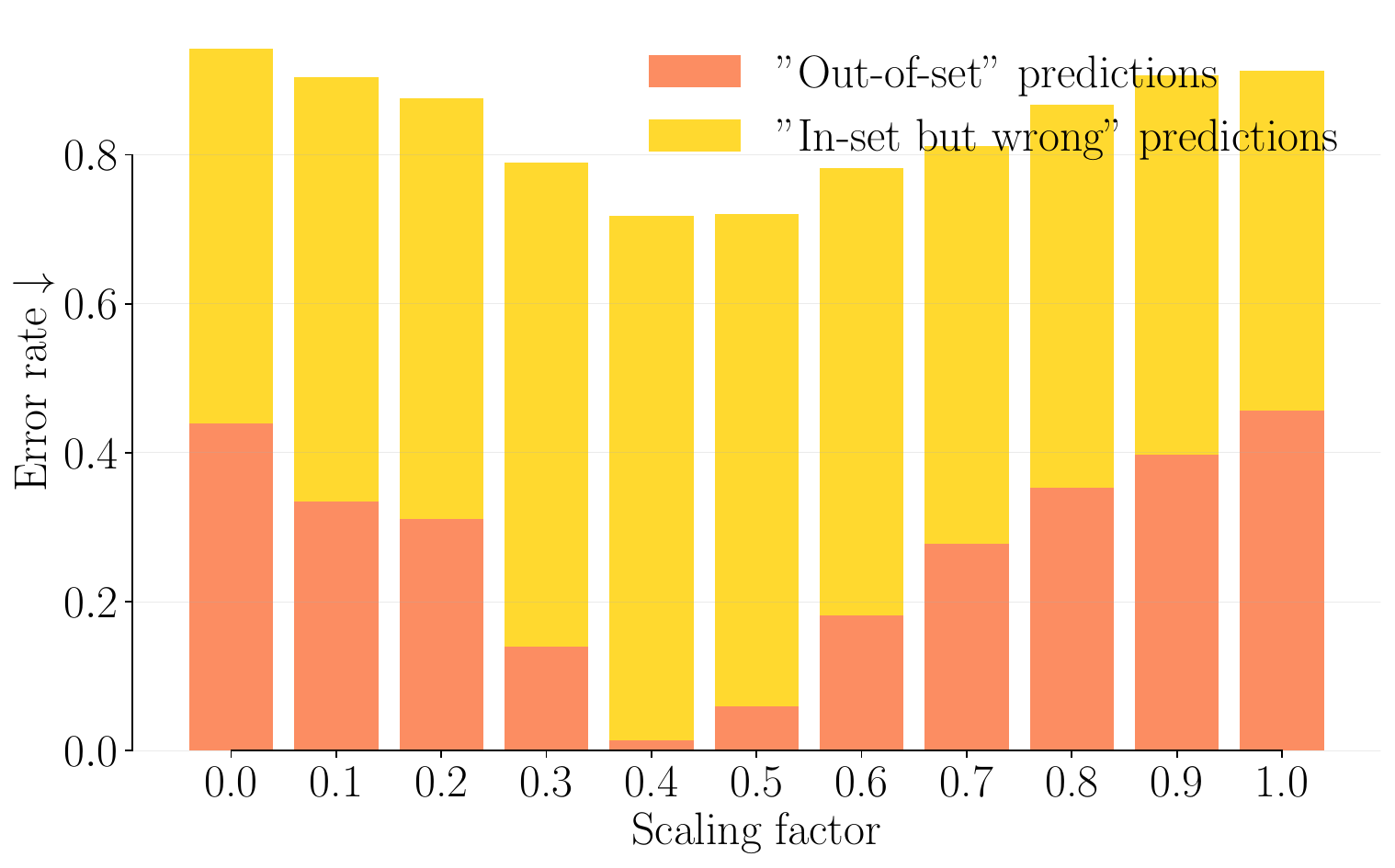}
  \caption{
    Error breakdown (lower is better) on \dataset{unseen-family-cmn} as a function of the merging coefficient $\alpha$.
  }
  \label{fig:unseen_cmn_family_error_rate_all}
\end{figure}

\subsection{Zero-Shot Generalization Task} \label{sec:zf-indiv}

\begin{figure}[t]
  \centering
  \begin{subfigure}[t]{0.47\textwidth}
    \vspace{0pt}
    \begin{promptbox}{Original Prompt}
    Is there only one bird in the audio, or more? Reply with `One' or `More'.
    \end{promptbox}\vspace{4pt}
    \begin{promptbox}{Reversed Classes Prompt}
    Is there more than one bird in the audio, or just one? Reply with `More' or `One'.
    \end{promptbox}\vspace{4pt}
    \begin{promptbox}{No Classes Prompt}
    How many birds are there in the audio?
    \end{promptbox}
    \caption{}
  \end{subfigure}\hfill
  \begin{subfigure}[t]{0.49\textwidth}
    \vspace{0pt}
    \includegraphics[width=\linewidth]{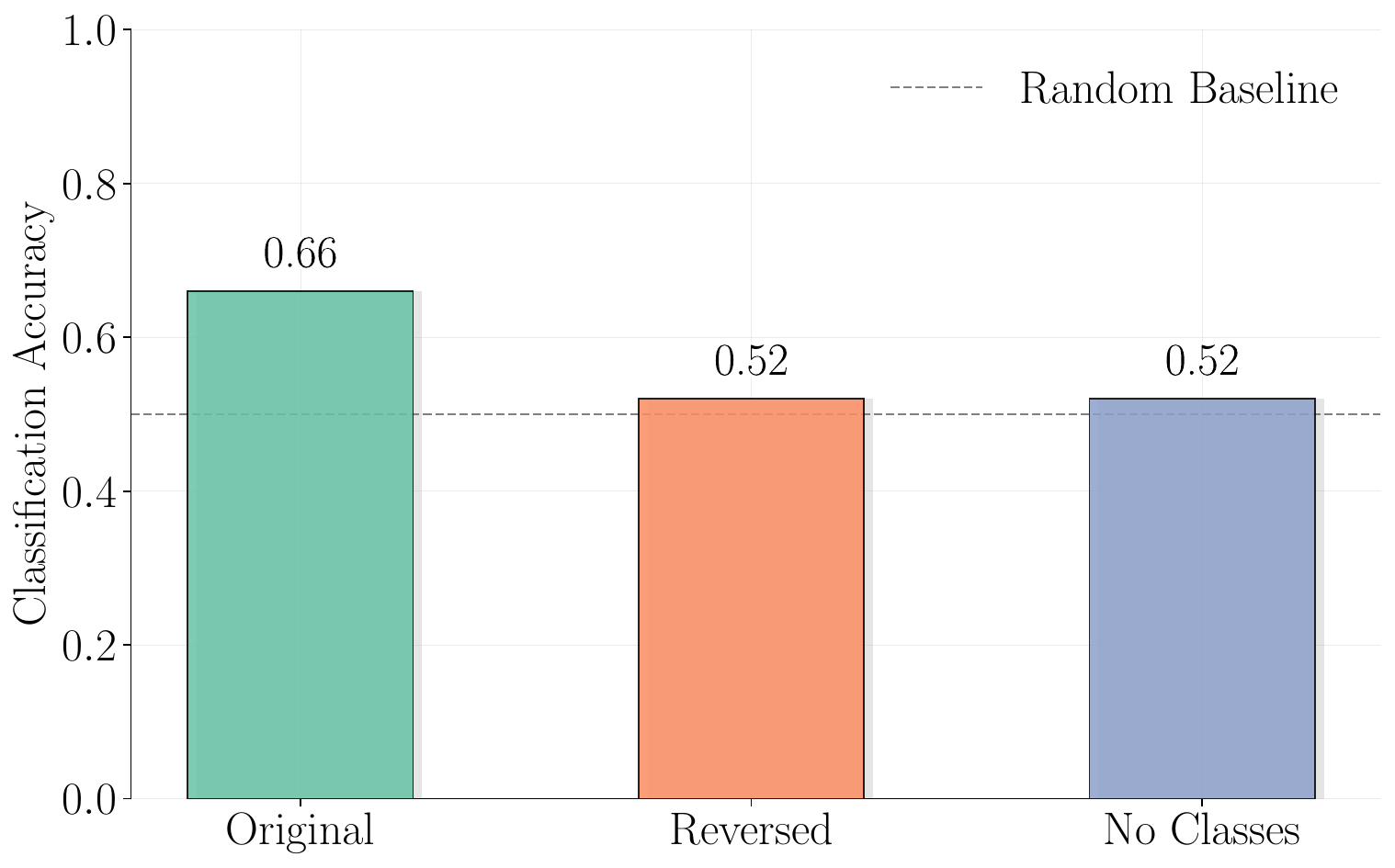}
    \caption{}
  \end{subfigure}
  \vspace{-4pt}
  \caption{
    \textbf{Classification accuracy for different prompt types on \dataset{zf-indiv}.}
    (a) Exact wording of the three evaluated prompts.
    (b) Accuracy of \model{NatureLM-audio} on \dataset{zf-indiv}.
    Accuracy is above random for the original prompt from~\citet{robinson2025naturelm}, but drops to near-random when the prompt is slightly reworded.
  }
  \label{fig:zf_indiv}
\end{figure}

In~\citet{robinson2025naturelm}, zero-shot generalization was evaluated on the \dataset{zf-indiv} dataset~\citep{elie2015zfindiv}, part of the \dataset{BEANS-Zero} benchmark~\citep{hagiwara2023beans}, as it was excluded from the model's training set. \dataset{zf-indiv} tests the ability to infer the number of Zebra Finch (\emph{Taeniopygia guttata}) individuals in an audio recording, making it a particularly relevant downstream task.

With the original prompt from~\citet{robinson2025naturelm} (\cref{fig:zf_indiv}), \model{NatureLM-audio} achieves $0.66$ accuracy (random baseline: $0.5$), indicating partial generalization to this unseen task. However, reversing the class name order or removing explicit class labels (see~\cref{fig:zf_indiv} for the exact prompts), reduces accuracy to $0.52$, effectively random performance.

These results suggest that the higher-than-random performance reported in~\citet{robinson2025naturelm} is sensitive to prompt formulation. While the original result remains valid for the tested prompt, our findings indicate that the apparent zero-shot generalization may partly stem from prompt-specific biases rather than task understanding.

\subsubsection{Few-Shot In-Context Learning}
\begin{figure}[t]
  \centering
  \begin{subfigure}[t]{0.47\textwidth}
    \vspace{0pt}%
    \begin{promptbox}{In-Context Learning Prompt (One Shot)}
        Identify the common name for the focal species in the audio. 
        Output exactly one of: Dall's Porpoise, Spotted Elachura
        \\ \\
        Audio: <Audio><SampleAudioHere></Audio>  \\
        Label: Dall's Porpoise
        \\ \\
        Audio: <Audio><SampleAudioHere></Audio> \\
        Label: Spotted Elachura
        \\ \\
        Audio: <Audio><TestAudioHere></Audio> \\
        Label: 
    \end{promptbox}
  \end{subfigure}\hfill
  \begin{subfigure}[t]{0.49\textwidth}
    \vspace{0pt}%
    \includegraphics[width=\textwidth]{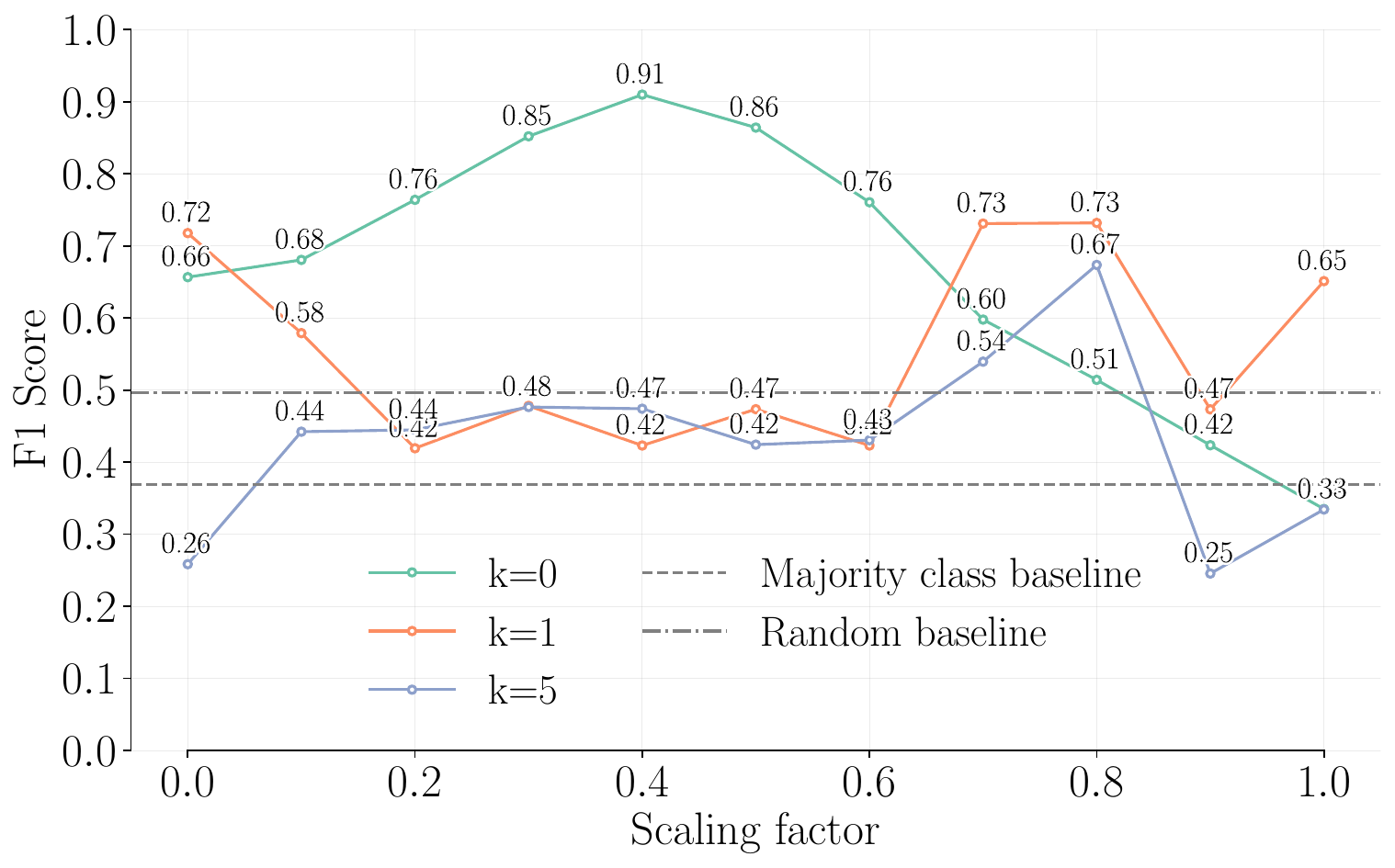}
  \end{subfigure}
  \vspace{-4pt}
  \caption{
    Example prompt and performance for few-shot in-context classification on \dataset{unseen-family-cmn} when varying the merging coefficient $\alpha$.
    The prompt (left) illustrates how few-shot examples are formatted using audio tokens, while the plot (right) reports the resulting F1-scores for $k \in {0, 1, 5}$. Adding in-context examples does not consistently improve performance and can lead to noisier behavior across $\alpha$ values.
  }
  \label{fig:in_context_learning}
\end{figure}

Given the merged model’s demonstrated ability to follow prompts, we further investigated whether it could also benefit from few-shot in-context examples, a setting particularly relevant to bioacoustics, where practitioners often have access to only a handful of labeled recordings for new species, habitats, or acoustic conditions. Specifically, we evaluated a binary version of the task by directly injecting the audio tokens obtained from the BEATs + Q-Former component of \naturelm into the prompt. We tested prompts containing no examples ($k=0$, identical to the Closed-Set Classification setup in~\cref{fig:unseen_cmn_family_appendix}), one example per class ($k=1$), and five examples per class ($k=5$). The prompts and results are shown in~\cref{fig:in_context_learning}. Across all $\alpha$ values, in-context examples did not yield consistent gains. Instead, performance often degraded or fluctuated unpredictably, with higher $k$ leading to noisier behavior. This trend mirrors recent findings showing that multimodal large language models frequently fail to benefit from few-shot multimodal prompting, and in some cases (e.g., Gemini 2.5 Pro) such context can even harm detection accuracy~\cite{robicheaux2025roboflow100vl}. The effect is likely exacerbated by the fact that neither \naturelm nor the base \llama model were trained on multi-audio prompts, making setups like~\cref{fig:in_context_learning} substantially out-of-distribution, even for interpolated $\alpha$ values.

\subsubsection{Compute Resources}
All experiments were conducted on one NVIDIA A100 GPU (40 GB), using 8 CPU cores and 32 GB of RAM, requiring 300 GB of disk space, primarily for storing the \dataset{BEANS-Zero} benchmark. However, since only a fraction of the benchmark was used, the actual memory footprint could be substantially smaller.

\newpage

\end{document}